%% file: main_arxiv_final.tex
\documentclass[journal]{IEEEtran}

\usepackage[cmex10]{amsmath}
\usepackage{amssymb}
\usepackage[mathscr]{eucal}
\usepackage{cite}
\usepackage{amsbsy}
\usepackage{esint}
\usepackage{tikz}

\input definition.tex

\ifCLASSINFOpdf

\else

\fi

\usepackage{algorithmic}

\usepackage{array}

\usepackage{fixltx2e}

\usepackage{stfloats}

\usepackage{url}

\begin{document}

\title{Neural Representation-Based Method for Metal-induced Artifact Reduction in Dental CBCT Imaging}


\author{\IEEEauthorblockN{Hyoung Suk Park, Kiwan Jeon, and Jin Keun Seo, Member, IEEE}\\

\thanks{Hyoung Suk Park and Kiwan Jeon are with the National Institute for Mathematical Sciences, Daejeon, 34047, Republic of Korea (e-mail: hspark@nims.re.kr; jeonkiwan@nims.re.kr)}
\thanks{Jin Keun Seo is with the School of Mathematics and Computing (Computational Science and Engineering), Yonsei University, Seoul, 03722, Republic of Korea (e-mail: seoj@yonsei.ac.kr)}
\thanks{Manuscript received XXX; revised  XXX. Corresponding author: J. K. Seo (email: seoj@yonsei.ac.kr).}}

\markboth{IEEE}%
{Park \MakeLowercase{\textit{et al.}}: }

\IEEEtitleabstractindextext{%
\begin{abstract}
This study introduces a novel reconstruction method for dental cone-beam computed tomography (CBCT), focusing on effectively reducing metal-induced artifacts commonly encountered in the presence of prevalent metallic implants. Despite significant progress in metal artifact reduction techniques, challenges persist owing to the intricate physical interactions between polychromatic X-ray beams and metal objects, which are further compounded by the additional effects associated with metal-tooth interactions and factors specific to the dental CBCT data environment. To overcome these limitations, we propose an implicit neural network that generates two distinct and informative tomographic images. One image represents the monochromatic attenuation distribution at a specific energy level, whereas the other captures the nonlinear beam-hardening factor resulting from the polychromatic nature of X-ray beams. In contrast to existing CT reconstruction techniques, the proposed method relies exclusively on the Beer--Lambert law, effectively preventing the generation of metal-induced artifacts during the backprojection process commonly implemented in conventional methods. Extensive experimental evaluations demonstrate that the proposed method effectively reduces metal artifacts while providing high-quality image reconstructions, thus emphasizing the significance of the second image in capturing the nonlinear beam-hardening factor.
\end{abstract}

}

\maketitle

\IEEEdisplaynontitleabstractindextext

\IEEEpeerreviewmaketitle

\section{Introduction}
Metal artifact reduction (MAR) in dental cone-beam computed tomography (CBCT) is challenging owing to the prevalence of metallic implants in patients. Multiple metallic objects, such as dental implants, in the scanned region can result in severe computed tomography (CT) image artifacts owing to the complex physical interactions between the polychromatic X-ray beams and metal objects. However, despite significant progress in MAR methods over the past four decades, existing approaches have shown limited performance in effectively reducing metal artifacts in dental CBCT environments, where multiple metal inserts occupy a significant area.

Dental CBCT has gained popularity as a cost-effective and low-radiation alternative to multidetector CT (MDCT) in dental clinics. However, a significant drawback of it is that its inverse problem is more challenging compared with MDCT. Specifically, it poses a highly complex and nonlinear challenge, primarily attributed to multiple factors, including intricate metal-bone and metal-tooth interactions, photon starvation, field-of-view truncation, offset detector, and scattering.
Metal-induced artifacts stem from the mismatch between the forward models employed in conventional reconstruction algorithms (such as filtered backprojection (FBP) \cite{Bracewell1967} and Feldkamp-Davis-Kress (FDK) \cite{Feldkamp1984}), and the polychromatic nature of X-ray beams. X-ray beams in dental CBCT comprise photons with energies ranging from minimum (e.g., 0 keV) to peak energy (e.g., between 60 and 120 keV)\cite{Pauwels2014}. However, these conventional algorithms overlook the polychromatic nature of X-ray beams, thus leading to a discrepancy between the sinogram data and the range space of the forward operator, such as the Radon transform. This discrepancy can result in widespread artifacts in the reconstructed image; the reconstruction process aims to minimize the discrepancy between the forward projection of the image and measured sinogram.

Over the past four decades, numerous methods for MAR have been developed, including projection-based methods \cite{Abdoli2010,Kalender1987,Lewitt1978,Meyer2010,Park2013,Roeske2003,Zhao2000}, iterative reconstruction methods \cite{DeMan2001,Elbakri2002,Menvielle2005,OSullivan2007,Wang1996}, dual-energy CT methods \cite{Alvarez1976,Lehmann1981,Yu2012}, and photon counting methods \cite{Layer2023,Patzer2023}.
Projection-based methods may encounter difficulties in correcting distorted data, particularly when metal objects are large or complex.
Iterative methods can achieve superior results compared with projection-based methods; however, they have limitations in accurately modeling complex interactions between X-rays and metal objects.
Dual-energy methods improve the accuracy of material identification and artifact reduction; however, they require specialized hardware or software, and an increased radiation dose.
Photon counting is a promising technology that has recently gained attention for its potential application in MAR \cite{Layer2023,Patzer2023}. However, it may not be suitable for dental CBCT because of the high cost of photon-counting detectors. Recently, deep learning algorithms have been widely utilized for MAR in X-ray CT and can be roughly classified into three categories: image-domain learning \cite{Gjesteby2017,Nakao2020,Zhang2018-1}, projection-domain learning \cite{Park2018}, and dual-domain learning \cite{Lin2019,Zhang2020}. The abovementioned methods require numerous paired metal-affected and metal-free CT scans for network training. However, obtaining paired datasets in clinical practice remains challenging. Furthermore, the performance of deep learning methods can considerably degrade when applied to CT scans acquired under acquisition conditions or CT scanners that differ from those used for training.

\begin{figure*}[!t]
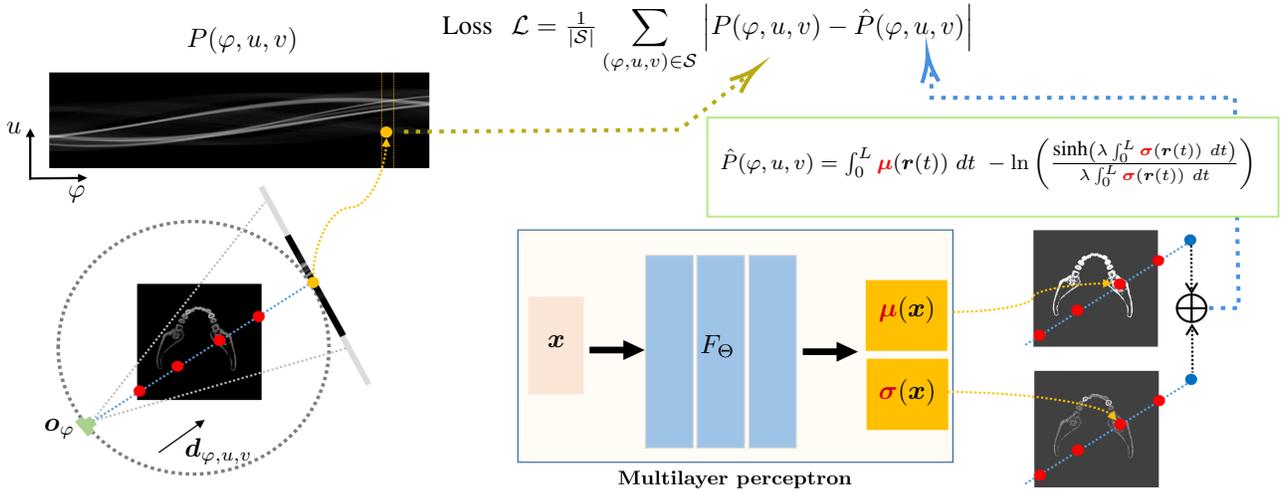

\centering
\input NeRF-CT.tex
\caption{Schematic diagram of the proposed method for metal artifact reduction (MAR) in dental cone-beam computed tomography (CBCT). The key aspect of the proposed method is that it differs from existing CT reconstruction techniques in that it exclusively relies on the multilayer perceptron and the formula of $\hat{\text{P}}$}.
\label{fig-main}
\end{figure*}

To address the intricate challenge of MAR in dental CBCT, we thoroughly investigated the limitations of conventional methods, such as FBP and FDK algorithms. Recognizing the need for an innovative approach that circumvents the backprojection process commonly used in these methods and its tendency to generate metal-induced artifacts, we proposed a novel MAR algorithm. Recently, neural radiance fields (NeRFs) \cite{Mildenhall2021} in computer vision have demonstrated considerable potential for representing 3D scenes from 2D camera data using deep neural networks. Inspired by this, we proposed a CT reconstruction method that utilizes the inherent capabilities of neural representations to generate two distinct, informative tomographic images. One image represents the monochromatic attenuation distribution at a specific energy level, whereas the other captures the nonlinear beam-hardening factor stemming from the polychromatic nature of X-ray beams. In contrast to the existing CT reconstruction techniques, the proposed method exclusively relies on the Beer--Lambert law, effectively preventing the generation of metal-induced artifacts during the backprojection process commonly employed in conventional methods. Figure \ref{fig-main} shows the schematic diagram of the proposed method.

The efficacy of the proposed method was assessed through evaluations of realistic simulated and phantom experiment datasets. The results demonstrated increased efficiency in reducing metal artifacts while preserving the morphological structures around metallic objects. Furthermore, the proposed method offers promising performance even with photon starvation.

\section{Mathematical framework}
In dental CBCT, a cone-shaped X-ray beam is directed through a patient's head while they are positioned between an X-ray source and a flat-panel detector housed in a gantry. The gantry is rotated to allow the X-ray beam to pass through the patient's head from various angles. During this process, a planar detector acquires the CBCT projection data denoted as $\text{P}(\varphi, u, v)$, where $\varphi\in [0,2\pi)$ represents the projection angle and $(u,v)$ represents the position of the planar detector. The position is scaled using the ratio of the distance between the X-ray source and detector plane to the distance between the source and rotation axes.

The sinogram $\text{P}$ acquired from low-dose dental CBCT can be described by the expression:
\begin{equation} \label{pfull}
	\text{P} =  \mathcal S_{\mbox{\scriptsize truncation}}( \text{P}_{\mbox{\scriptsize full}}),
\end{equation}
where  $\text{P}_{\mbox{\scriptsize full}}$ denotes the corresponding sinogram acquired using a wide-detector CBCT without any offset, thus providing the entire information for a sinogram; and $\mathcal S_{\mbox{\scriptsize truncation}}$ represents the truncation operator determined by the size and offset configuration of the detector.

The main objective here is using the truncated data $\text{P}$ to reconstruct a scalar value $\mu(\x)$ that represents the attenuation coefficient at a fixed energy level $E_0$ and for a specific position $\x=(x,y,z)$ in world coordinates. Under the idealized monochromatic assumption, a linear X-ray transform exists, denoted by $\mathcal T_{\text{\tiny fw}}$, such as the Radon or cone-beam transforms, which maps the CT image to the projection data as follows.
\begin{equation}\label{Forward1}
\text{P} = \mathcal T_{\text{\tiny fw}}~ \mu.
\end{equation}
However, this monochromatic model is inaccurate because the X-ray beams used in these scans consist of photons with a range of energies. Thus, the X-ray attenuation coefficient distribution, denoted by $\mu_E (\x)$, varies with the position $\x$ and photon energy level $E$.
Consider the path of the X-ray beam from the source position ${\bf o}_{\varphi}$ to the detector position $\x_{\varphi, u,v}$ in the world coordinates. Owing to the polychromatic nature of X-ray beams, the projection data $\text{P}(\varphi, u, v)$ follow the Lambert-Beer law \cite{Beer1852,Lambert1892}.
\begin{align}\label{Lambert-Beer-law}
\text{P}(\varphi, u, v)
=-\ln\left(\int_{E_{\text{min}}}^{E_{\text{max}}}\eta(E)\exp \left(-\int_{\ell_{\varphi, u,v}} \mu_E ds\right)dE\right),
\end{align}
where $\int_{\ell_{\varphi, u,v}} \mu_E ds$ is the line integral of $\mu_E$ over the ray $\ell_{\varphi, u,v}$ joining the source position ${\bf o}_\varphi$ and detector position $\x_{\varphi, u,v}$; and $\eta(E)$ represents the fractional energy at photon energy $E$ in the spectrum of the X-ray source \cite{Herman1983}, with its support being the interval $[E_{\text{min}},E_{\text{max}}]$, and $\int_{\Bbb R} \eta(E) dE=1$.

\subsection{Inherent drawbacks of methods using FBP or FDK}
To solve the ill-posed problem, a regularized least squares method of the following form can be used:
\begin{equation}\label{leastSquare1}
	\mu_*=\underset{\mu}{\mbox{argmin}}   \|\text{P} - \mathcal T_{\text{\tiny fw}}~ \mu\|_{\ell_2}^2 +\gamma \text{Reg}(\mu),
\end{equation}
where $\text{Reg}(\mu)$ is a regularization term constraining prior knowledge of artifact-free and noise-free CBCT images; $\|\cdot\|_{\ell_2}$ denotes the standard Euclidean norm; and $\gamma$ is the regularization parameter controlling the trade-off between the fidelity term and regularity.

\begin{figure*}[!t]
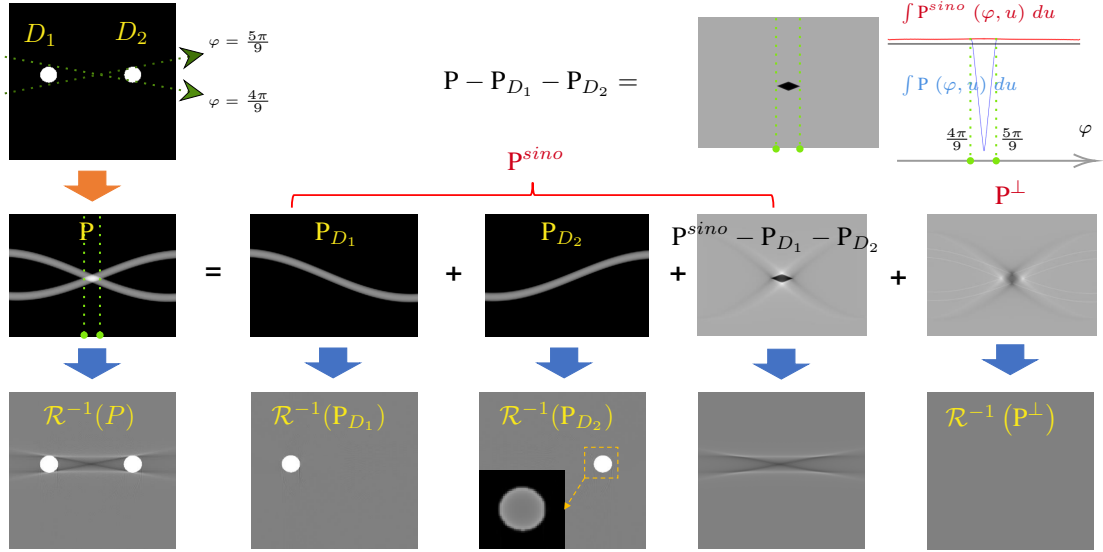

\centering
\input MAR-CT.tex
\caption{Characterization of metal-induced artifacts observed in a two disk-shaped phantom. The projection data $\text{P}$ exhibits local inconsistency $\text{P}-(\text{P}_{D_1} + \text{P}_{D_2})$, which leads to the emergence of global artifacts, such as streaking and shadowing, in the reconstructed CT image. These artifacts manifest during fitting $\text{P}$ onto $\text{P}^{sino}$ in the range space $\mathcal H^{sino}$ when using the filtered backprojection (FBP) method. In the bottom figures, the symbol $\mathcal R^{-1}$ represents the FBP operation. The middle image in the bottom row emphasizes cupping artifacts in the disk region.}
\label{fig-mar-ct}
\end{figure*}

The linear operator can be expressed as follows:
\begin{equation}\label{linear-pojection}
 \mathcal T_{\text{\tiny fw}}: \mu \in \R^{V}	\mapsto\text{P}\in \R^{S\times D}
\end{equation}
where $V$ denotes the numebr of voxels in the CBCT images, $S$ denotes the number of views, and $D$ denotes the number of detector cells. According to the Hiblert projection theorem, the Hilbert space $\mathcal H=\R^{S\times D} $ can be decomposed as:
\begin{equation}\label{Hilbert-pojection}
	\mathcal H= \mathcal H^{sino} \oplus \mathcal H^{\perp}
\end{equation}
where $\mathcal H{sino}=\{\mathcal T_{\text{\tiny fw}}\mu : \mu \in \R^{V}\}$ is the range space,  $\mathcal H^{\perp}$ is its orthogonal complement, and $\bigoplus$ denotes the orthogonal direct sum.
Hence,  $\text{P}$ can be decomposed into
\begin{equation}\label{Hilbert-pojection2}
	\text{P}=\text{P}^{sino} + \text{P}^{\perp}
\end{equation}
where $\text{P}^{sino}\in \mathcal H^{sino}$ and $\text{P}^{\perp}\in \mathcal H^{\perp}$.
Thus, the problem is equivalent to:
\begin{equation}\label{leastSquare2}
	\mu_*=\underset{\mu}{\mbox{argmin}}   \|\text{P}-\text{P}^{\perp}  - \mathcal T_{\text{\tiny fw}}~ \mu\|_{\ell_2}^2 +\gamma \text{Reg}(\mu).
\end{equation}
Note that $\mathcal T_{\text{\tiny fw}}$ maps an arbitrary single voxel image to the corresponding sinusoidal curve in the sinogram space $\mathcal H$. Hence, any single-pixel mismatch in $\text{P}$ leads to a sinusoidal global change $\text{P}^{\perp}$ when inputting data into the range space $\mathcal H^{sino}$. Thus, attempting a local mismatch in $\text{P}$ is highly desirable; however, this is not possible within the above least-squares framework. Global matching of $\text{P}$ by subtracting $\text{P}^{\perp}$ produces streaking or shadowing artifacts (see Fig. \ref{fig-mar-ct}).

To provide a rigorous explanation of cupping and streaking artifacts for metallic objects in CT imaging, we focus on the fan-beam CT model, where we restrict $\text{P}(\varphi, u,0)$ to detector position $v=0$. We can then represent $\mathcal{T}_{\text{\tiny fw}}$ as a composition of the Radon transform and the data-filtering operator that converts the fan-beam projection data into a parallel beam sinogram. To explain how $\text{P}^\perp$ destroys the global structure of $\text{P}$, we examined a simplified model comprising two disk-shaped metallic objects, as shown in Fig. \ref{fig-mar-ct}. Specifically, the desired ideal CT image can be represented as $\mu=c\chi_{D_1\cup D_2}$ (where $c$ is a constant, $D_1$ and $D_2$ are disks of equal radius, and $\chi_D$ denotes the characteristic function of region $D$), by assigning it a value of one inside $D$ and zero otherwise.
To analyze the projection data $\text{P}$, we introduce $\text{P}_{D_1}$ to denote the projection data solely related to $D_1$, and $\text{P}_{D_2}$ for $D_2$. Interestingly, $\text{P}_{D_1}$ and $\text{P}_{D_2}$ lie within the range space but yield cupping artifacts \cite{Park2015,Park2017}. Therefore, $\text{P}_{D_1}$ and $\text{P}_{D_2}$ are consistent and $\text{P}_{D_1}^\perp=0=\text{P}_{D_2}^\perp$. By contrast, $\text{P}$ exhibits inconsistency, thus leading to $\text{P}^\perp\neq 0$, as shown in Fig. \ref{fig-mar-ct}. Here, $\text{P}^\perp$ was computed as $\text{P}^\perp=\text{P}-\mR\mR^{-1}\text{P}$, where $\mR$ and $\mR^{-1}$ denote the Radon transform and FBP operators, respectively. Consider a scenario in which an X-ray beam passes through both disks within a projection angle range of $4\pi/9$ to $5\pi/9$. Thus, $\text{P}(\phi,u)\neq \text{P}_{D_1}(\phi,u)+ \text{P}_{D_2}(\phi,u)$ for $\phi$ within the range $[4\pi/9, 5\pi/9]$, whereas $\text{P}(\phi,u)= \text{P}_{D_1}(\phi,u)+ \text{P}_{D_2}(\phi,u)$ holds true for $\phi$ outside this interval.
Based on the sinogram consistency condition for $\text{P}^{sino}$, it follows that for all $\phi \in [4\pi/9, 5\pi/9]$ and $\phi' \notin [4\pi/9, 5\pi/9]$,
\begin{equation}\label{ortho}\int (\text{P}(\phi, u)-\text{P}^\perp(\phi,u))du=\int (\text{P}(\phi', u)-\text{P}^\perp(\phi', u))du.\end{equation}
This indicates that $\text{P}^\perp$ corrects specific regions and affects the global structure of $\text{P}$ in a broader sense. As shown in Fig. \ref{fig-mar-ct}, $\text{P}^\perp$, used for rectifying the mismatch, has a broad impact on the entire sinogram, thus leading to the deterioration of its global structure and introduction of streaking and shadowing artifacts. Existing methods that use the backprojection process cannot offer localized correction solely to $\text{P}$ within the projection angle range of $[4\pi/9, 5\pi/9]$ without influencing other segments of the sinogram $\text{P}$. Consequently, novel methods that address this issue and provide localized corrections specifically to the relevant regions of the sinogram while avoiding adverse impact on other portions must be urgently developed.

\subsection{Fundamental structure of global artifacts caused by sinogram inconsistency}
This section investigates the structure of artifacts caused by a sinogram inconsistency. Assume that $\text{P}$ has a local mismatch $\text{P}^{\mbox{\tiny mismatch}}$ whose support occupies a small area in the sinogram space. The corrected sinogram $\text{P}-\text{P}^{\mbox{\tiny mismatch}}$ is in the range space such that $\mu_*$ exists, where $\mathcal T_{\text{\tiny fw}}~ \mu_*=\text{P}-\text{P}^{\mbox{\tiny mismatch}}$.

To simplify notation, we will denote a position $(\varphi, u, v)$ in sinogram space as $\xi = (\varphi, u, v)$.
Let us consider the scenario where a sinogram mismatch occurs at a single point $\xi_0=(\varphi_0, u_0, v_0)$. If this mismatch is a Dirac function $\delta_{\xi_0}$, then the corresponding artifact can be represented as:
\begin{equation}\label{f-artifact1}
	\Gamma _{\xi_0}=\underset{\mu}{\mbox{argmin}}   \|\delta_{\xi_0} - \mathcal T_{\text{\tiny fw}}~ \mu\|_{\ell_2}^2
\end{equation}
Then, the artifacts caused by the sinogram inconsistency $\text{P}^\perp$ can be expressed as:
\begin{equation}\label{f-artifact2}
	\Upsilon(\x)=\int_{\Omega} \Gamma_{\xi}(\x) \text{P}^{\mbox{\tiny mismatch}}(\xi) d\xi
\end{equation}
where $\Omega$ is the support of $\text{P}^{\mbox{\tiny mismatch}}$.

\begin{remark} To understand metal-induced artifacts more intuitively, let us consider a simplified scenario of a bichromatic model with energies of 64 and 80 KeV and the fractional energy is described as $
\eta(E)=\frac{1}{2} \delta(E-64)+\frac{1}{2} \delta(E-80)$. We want to reconstruct an image that is a $3\times 3$ pixel matrix, which is represented as:
   $$
   \left(
   \begin{array}{ccc}
      \mu_{1,1} &   \mu_{1,2} & \mu_{1,3} \\
      \mu_{2,1} &   \mu_{2,2} & \mu_{2,3} \\
      \mu_{3,1} &   \mu_{3,2} & \mu_{3,3}\\
   \end{array}
   \right),
   $$
   where $\mu_{2,1}=\mu_{2,3}$ are metals and the rest are air. We hope that the reconstructed image should be of the form
\begin{equation}\label{sol}
   \left(
\begin{array}{ccc}
   0 &   0 & 0\\
   c &   0 &  c \\
   0 &   0 & 0\\
\end{array}
\right),
\end{equation}
for some constant $c$ associated with the attenuation coefficient of the metal. The attenuation coefficients of the metal are 64 at $E=64$ keV and 5 at $E=80$ keV. Assume that we have the projection data of three angles $\varphi=0,\frac{\pi}{4}, \frac{\pi}{2}$.
Then, the conventional CT reconstruction problem solves the following system.
\begin{equation}\label{RT}
\left\{\begin{array}{llll}
      \mu_{1,1} +   \mu_{2,1} + \mu_{3,1} &=   &\text{P}(0, 1)&=5.7  \\
   \mu_{1,2} +   \mu_{2,2} + \mu_{3,2} &=   &\text{P}(0, 2)&=0  \\
   \mu_{1,3} +   \mu_{2,3} + \mu_{3,3} &=   &\text{P}(0, 3)&=5.7 \\
  \mu_{2,1} +   \mu_{3,2}  &=   &\text{P}(\pi/4, 1)&=5.7\\
   \mu_{1,1} +    \mu_{2,2} +  \mu_{3,3} &=   &\text{P}(\pi/4, 2)&=0\\
   \mu_{1,2} +   \mu_{2,3} &=   &\text{P}(\pi/4, 3)&=5.7 \\
   \mu_{3,1} +   \mu_{3,2} + \mu_{3,3} &=   &\text{P}(\pi/2, 1)&=0\\
   \mu_{2,1} +   \mu_{2,2} + \mu_{2,3} &=   &\text{P}(\pi/2, 2)&=10.7\\
   \mu_{1,1} +   \mu_{1,2} + \mu_{1,3} &=   &\text{P}(\pi/2, 3)&=0 \\
   \end{array}
   \right.
\end{equation}
where 10.7 comes from
$
10.7 \approx  - \log (0.5\exp(-64\times2) + 0.5\exp(-5\times2))
$
 and 5.7 comes from $5.7\approx  - \log (0.5\exp(-64\times1) + 0.5\exp(-5\times1)) $.
The standard CT reconstruction algorithm is to find $\boldsymbol \mu_{\CT}$ such that
$$
\boldsymbol \mu_\CT = \underset{\boldsymbol \mu}{\mbox{argmin}}
\| A \boldsymbol \mu - \text{P} \|_{\ell_2}^2, $$
where $\A$ is the $9\times 9$ matrix corresponding to the Radon transform in \eqref{RT} and  $\mu$  can be understood as a vectorized version.  The reconstructed image using the formula $\mu_\CT =(\A^T\A)^{-1}\A^T \text{P}$ is given by $$
\left(
\begin{array}{ccc}
   -1.0 &   2.2 & 0.4\\
   6.8 &   2.5 &  6.3 \\
   0.2 &  -0.5 & 0.7\\
\end{array}
\right).
$$ Note that the reconstructed image $\boldsymbol{\mu}_{\text{CT}}$ significantly deviates from the true solution in \eqref{sol} owing to the backprojection process $\A^T\text{P}$.
This discrepancy can be attributed to the single mismatch observed in the 8th equation of \eqref{RT}, where $\text{P}(\pi/2, 2)=10.7\neq 2\times 5.7$.
\end{remark}

\begin{figure*}[!t]
\centering
\includegraphics[width=1\textwidth]{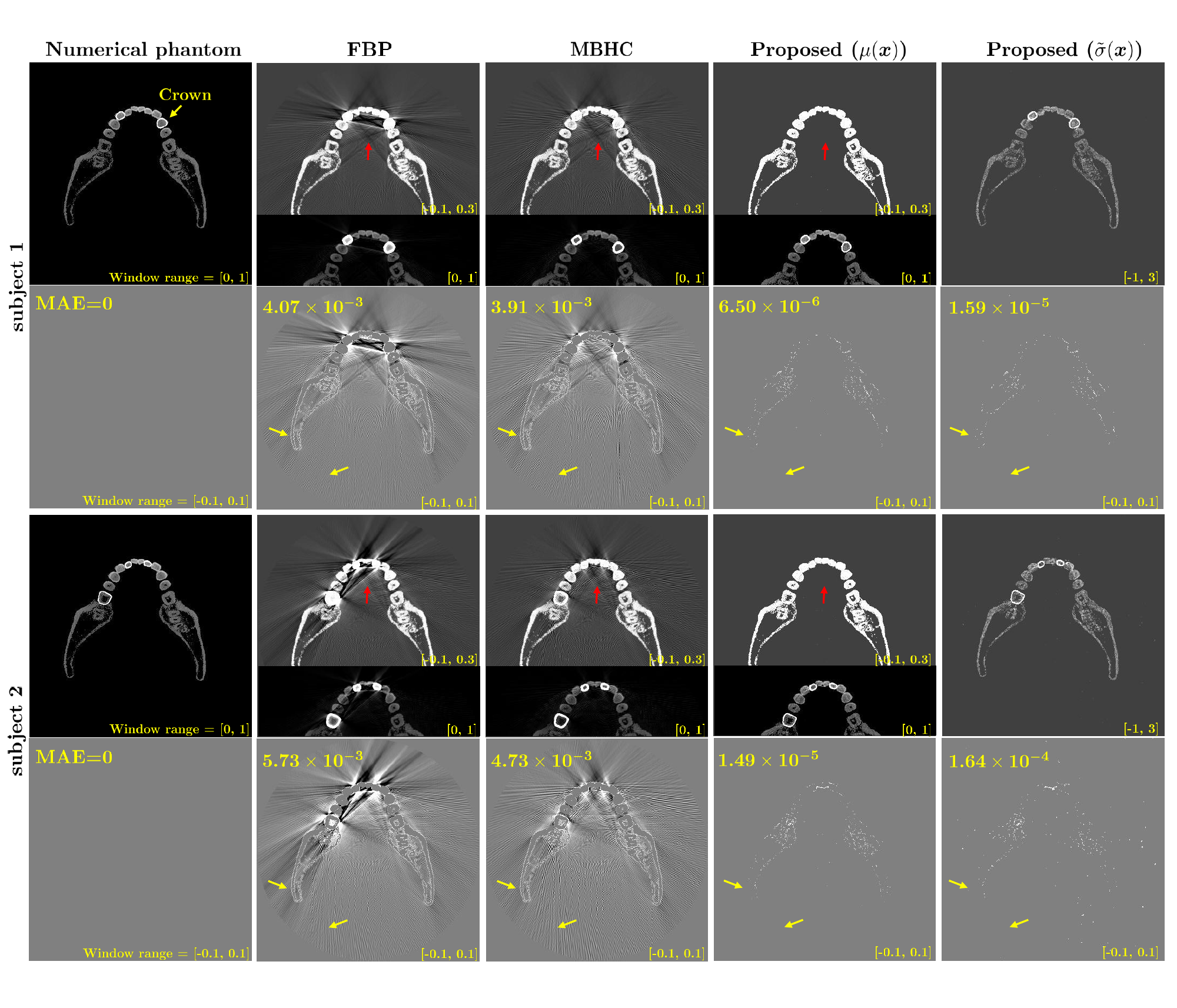}
\caption{Comparison of the reconstruction results for numerical phantom consisting of teeth, bone, and multiple crowns. The second and fourth rows show the background of the reconstructed CT images, which correspond to the air region without teeth, bone, and crowns.}
\label{fig-num-result}
\end{figure*}

\subsection{Implicit neural representation-based MAR}
Conventional CBCT reconstructions use a pixel or voxel-based approach to represent images; however, using this approach in low-dose dental CBCT is challenging owing to the large dimension of the solution space and inconsistent data in the presence of metal implants. To address these issues, it is crucial to incorporate an image prior that constrains the relationships between pixels based on underlying head anatomy. Although regularization techniques are commonly used for this purpose, their performance is limited because they lack global control between pixels.

By contrast, neural representations using multilayer perceptrons (MLPs) utilize implicit representations that can capture complex relationships between image pixels more efficiently. These representations enable a significant reduction in the dimensions of the solution space, thus offering a more efficient and accurate reconstruction with highly undersampled data.

Our approach to solving the inverse problem of dental CBCT is inspired by the recent success of NeRF in accurately representing 3D scenes derived from 2D camera data using a deep learning network. The proposed approach uses MLP to encode CT representations. The MLP takes a 3D point $\x=(x,y,z)$ as the input and outputs the attenuation coefficient $\mu(\x)=\mu(\x,E_0)$ and its energy-dependent beam-hardening factor $\sigma(\x):=\frac{\partial}{\partial E}\mu (\x, E_0)$.
\begin{equation}\label{MARNeRF}
   f_\Theta: \x \mapsto (\mu(\x), \sigma(\x)) .
\end{equation}
Instead of directly computing the attenuation coefficient $\mu(\x)$, we use $f_\Theta$ rather than the standard expression for $\mu$ because it provides a more concise representation of the CT image while producing the same $\mu(\x)$ as the standard expression. This compact implicit expression allows us to solve the inverse problem with highly undersampled data $\text{P}$.

To learn function $f_\Theta$, we minimize the difference between the measured data $\text{P}$ (ground truth) and the predicted data $\hat{\text{P}}$, generated using the output of $f_\Theta$. The loss function is defined as
\begin{equation}\label{linear-approximation3-0}
   \mathcal L = \f{1}{|\mS|}\sum_{ (\varphi,u,v) \in \mathcal S} |\hat{\text{P}}(\varphi,u,v) -\text{P}(\varphi,u,v)|,
\end{equation}
where $\mathcal S$ represents the set of X-rays that pass through the detector positions.

Next, we explain computing $\hat{\text{P}}(\varphi,u,v)$ from $f_\Theta$. Consider the X-ray path $\br(t)={\bf o}_{\varphi}+ t\bd_{\varphi,u,v}$, $t\in [0,L]$, where ${\bf o}_{\varphi}$ is the X-ray source position and $\bd_{\varphi,u,v}=(\sin\varphi,-\cos\varphi, \beta v)$ is a direction vector of the X-ray corresponding to the position $(\varphi, u, v)$ in the projection data $\text{P}$. This path is defined for $t\in [0,L]$, where $L$ is the path length.

We use $f_\Theta$ to compute $\mu (\br(t))$ and $\sigma (\br(t))$.
A careful analysis reveals that $\hat{\text{P}}(\varphi,u,v)$ can be approximately computed as follows.
\begin{equation}\label{linear-approximation3}
 \hat{\text{P}}(\varphi,u,v)= \int_0^L \mu(\br(t)) dt -\ln \left(\frac{ \mbox{sinh}(\lambda \int_0^L \sigma(\br(t))dt) }{\lambda \int_0^L \sigma(\br(t))dt} \right),
\end{equation}
where $\lambda>0$ is a constant depending on CBCT scanning system.

Now, we provide the proof of \eqref{linear-approximation3}.
From the Beer-Lambert law \eqref{Lambert-Beer-law}, we have
\begin{align}\label{linear-approximation4}
   \text{P}(\varphi,u,v) &= -\ln \int_{E_{\text{min}}}^{E_{\text{max}}} \eta(E)\exp\left[-\int_0^L \mu (\br(t), E_0) \right. \nonumber\\
   &\left. + (E-E_0)\frac{\partial}{\partial E}\mu (\br(t), E_0)dt\right ] dE,
\end{align}
where $E_0$ is a reference energy level and the partial derivative of the attenuation coefficient $\mu$ with respect to photon energy $E$ is evaluated at $E_0$. This expression leads to the following approximation.

\begin{align}\label{linear-approximation2}
  \text{P}(\varphi,u,v) & \approx \int_0^L \mu (\br(t)) dt \nonumber\\ &- \ln \int_{-1}^{1} \frac{1}{2}\exp\left[- \lambda s\int_0^L \sigma (\br(t))dt\right ] ds.
\end{align}
Direct computation of \eqref{linear-approximation2} yields \eqref{linear-approximation3}, which completes the proof.

In practice, accurately estimating the parameter $\lambda$ in $\hat{\text{P}}$ is challenging. Alternatively, for any constant $\hat{\lambda}$, $\hat{\text{P}}$ can be reformulated as follows.
\begin{align}\label{linear-approximation5}
 \hat{\text{P}}(\varphi,u,v)&= \int_0^L \mu(\br(t)) dt \nonumber\\ & -\ln \left(\frac{ \mbox{sinh}\left(\hat{\lambda} \left(\int_0^L \tilde{\sigma}(\br(t))dt +\eps\right)\right)}{\tilde{\lambda} \left(\int_0^L \tilde{\sigma}(\br(t))dt +\eps\right)}  \right),
\end{align}
where $\tilde{\sigma}$ is a scaled version of $\sigma$, expressed as  $\tilde{\sigma}= (\lambda\slash\tilde{\lambda})|\sigma|$. Based on this formulation, we train $f_{\Theta}$ to provide $(\mu, \tilde{\sigma})$ with a suitably selected $\tilde{\lambda}$. In eq. \eref{linear-approximation5}, to ensure training stability and avoid division by zero, we incorporate a small positive value $\epsilon > 0$ in the numerator and denominator of $\hat{\text{P}}$. In this study, we consistently set $\tilde{\lambda}$ to three, which demonstrates stable performance across our experiments. 

\subsection{Implicit Neural Representations with Sinusoidal Activations}
To enhance the ability of the network $f_\Theta$ to accurately model data with high frequency variations, the  $f_\Theta$ is designed with a sinusoidal activation function \cite{Sitzmann2020}:
\begin{align}
f_\Theta(\x) = \W_n\left(\psi_{n-1}\circ \psi_{n-2}\circ\cdots\circ\psi_0\right)(\x) + \bb_n,
\end{align}
where $\W_i$ and $\bb_i$ are the weight and bias at the $i^{th}$ layer of the network, respectively. Further, $\psi_i$ is the $i^{th}$ layer of the network and is expressed as:
\begin{align}
\psi_i (\x_i) = \sin(\W_i\x_i + \bb_i).
\end{align}
The sinusoidal activation function can better represent the function, its derivative, and Laplacian information compared with the positional encoding method \cite{Mildenhall2021,Tancik2020}, which applies a serious of sine and cosine transforms to the input coordinates $\x$.

In our experiments, $f_\Theta$ consisted of five fully connected layers in between input and output layers. Each fully connected layer comprises 128 nodes, whereas the input and output layers each consist of two nodes. The network weights were updated using the Adam optimizer \cite{Kingma2015} at a learning rate of $5\times10^{-4}$. The training process was terminated when the loss function value in \eref{linear-approximation3-0} fell below $5\times10^{-3}$ for the numerical simulation and $9\times10^{-3}$ for the phantom experiment. The training procedure is implemented using PyTorch \cite{Paszke2019} on a system equipped with two CPUs (Intel(R) Xeon Gold 6226R, 2.9 GHz) and a GPU (NVIDIA RTX 3090, 24GB). Training the network per 2D CT image took approximately 3--5 min.

\begin{figure*}[!t]
\centering
\includegraphics[width=0.9\textwidth]{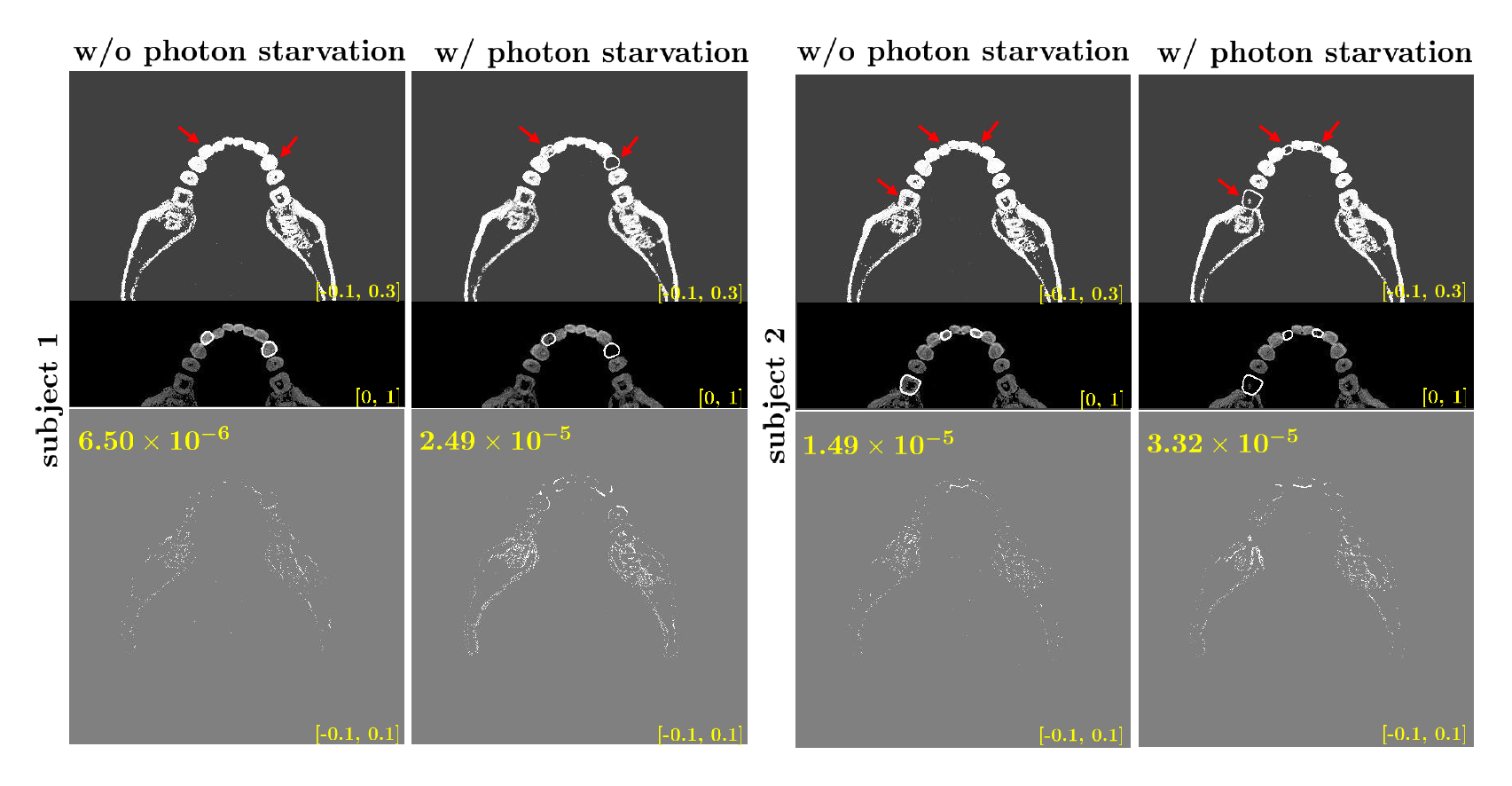}
\caption{Performance comparison of the proposed method for the photon starvation effect. In photon starvation (labeled as `w/ photon starvation'), the proposed neural network is trained using a subset of X-rays that passes only through the teeth and bone. The crowns segmented from the FBP image are additionally added to the reconstructed image.}
\label{fig-num-result_ps}
\end{figure*}

\begin{figure*}[!t]
\centering
\includegraphics[width=1\textwidth]{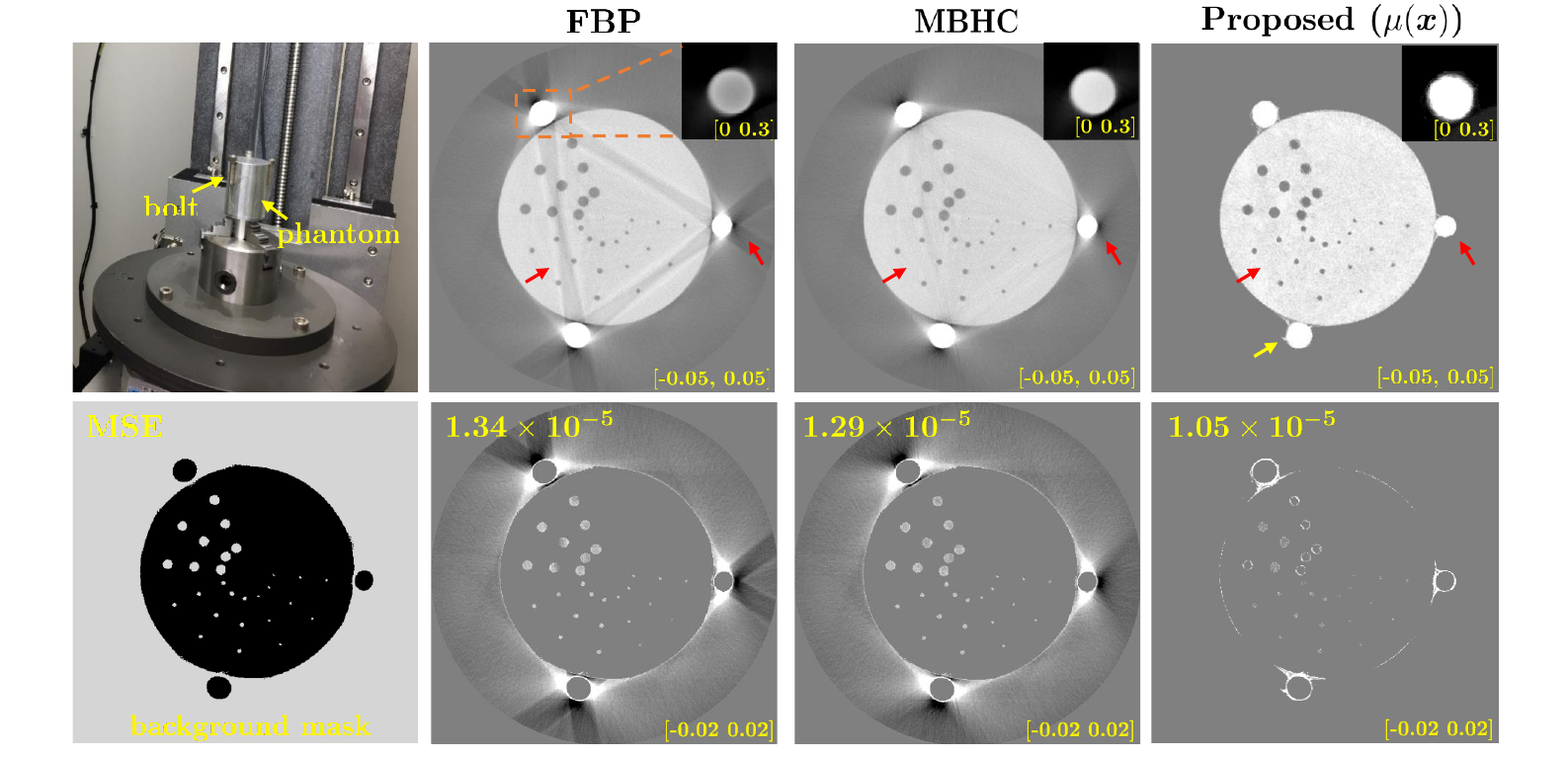}
\caption{Comparison of the reconstruction results for resolution phantom with three metallic bolts.}
\label{fig-phan_exp}
\end{figure*}

\begin{remark} In the field of medical tomographic image reconstruction, the conventional approach typically relies on a pixel-based (or voxel-based) representation, where each pixel corresponds to a dimension in the solution space. This process is particularly challenging when faced with a highly ill-posed reconstruction problem. The objective is to explore a vast solution space to and identify a single point that accurately represents the desired image. However, due to the inherent high-resolution nature of medical imaging, the solution space is primarily dominated by noise-like images, while practical solutions that resemble actual medical images occupy an incredibly small fraction, practically negligible in terms of probability. To mitigate these difficulties, researchers have developed various regularization techniques over the past several decades. These techniques aim to impose strong constraints on the solution to improve the reconstruction outcomes. However, these regularization methods often exhibit limited performance and loss of intricate details in the images. Implicit neural representation through MLPs shows promise for overcoming these limitations by optimizing its parameters to effectively search for the most appropriate solution within its architecture, thus offering a potential breakthrough in the field.
\end{remark}

\section{Results}
\subsection{Numerical Simulation}
To assess the effectiveness of the proposed method, we conducted a performance evaluation using a 2D numerical phantom. The phantom consisted of teeth, bones, and multiple crowns, as shown in Fig. \ref{fig-num-result}. Individual teeth were segmented as in \cite{Jang2021}, and a virtual crown was generated using dilation and erosion functions as in \cite{Hyun2022,Park2022}. The geometries of the teeth and bones were obtained by manually segmenting a real CBCT image.

The generated teeth, bone, and crowns were projected based on the X-ray polychromatic model in \eref{Lambert-Beer-law}. Here, we utilized the attenuation coefficients provided by the National Institute of Standards and Technology \cite{Hubbell1995} along with the energy spectrum $\eta(E)$ generated using the Spektr software \cite{Punnoose2016} at a tube voltage of 100 kVp. The crowns were composed of titanium. Additionally, we added Poisson and electric noise to the projection data, and disregarding other factors, such as photon starvation, scattering, and nonlinear partial volume effects. All $413\times 413$ was reconstructed with a pixel size of $0.4~\text{mm} \times 0.4~\text{mm}$.

We compared the performance of the proposed method with that of FBP and metal beam hardening correction (MBHC) methods. The FBP images were reconstructed using a standard Ram-Lak filter. In the MBHC method, the beam-hardening artifacts caused by metals were addressed using the following correction formula:
\begin{align}\label{bhc_formula}
  \phi_{D,\kappa}(\x) = -\mR^{-1}\left[\ln\left(\f{\sinh(\kappa\mR\chi_D)}{\kappa\mR\chi_D}\right)\right](\x).
\end{align}
In this method, we segmented the metal region $D$ using a simple thresholding approach. The optimal parameter $\kappa$ was chosen as $\kappa = 3$ based on Equation (17) in \cite{Park2015}. Based on \eref{linear-approximation2}, the two parameters $\kappa$ in \eref{bhc_formula} and $\lambda$ in \eref{linear-approximation3} are related as follows.
$$\kappa = -\alpha\lambda,$$
where the parameter $\alpha$ is defined as $\alpha=\f{\p}{\p E}\mu(\x,\E_0), \x\in D$.

Fig. \ref{fig-num-result} compares the reconstruction results for the numerical phantom. The second and third columns show CT images reconstructed using FBP and MBHC, respectively, whereas the fourth and fifth columns show that of the proposed method. The second and fourth rows show the backgrounds of the reconstructed CT images, which correspond to the air region (i.e., $\mu(\x)=0, \sigma(\x)=0$) without teeth, bones, and crowns. The mean absolute error (MAE) was computed and is listed in the upper-left corner of each background image.

Evidently, the FBP image suffered from severe streaking and shadow artifacts, primarily owing to the beam hardening effect caused by the crowns and teeth. The MBHC method reduced the metal beam-hardening artifacts between crowns in the FBP image. However, the artifacts from the interaction between the crowns and teeth remained (red arrows in the third column) because the metal beam-hardening corrector $\phi_{D,\kappa}$ only addresses the interactions between crowns.

The proposed method successfully reconstructed the attenuation ($\mu$) and its scaled energy dependent beam hardening factor ($\tilde{\sigma}$) images. The proposed method, as opposed to FBP and MBHC methods, successfully reduced the streaking and shadowing artifacts in the reconstructed images. Notably, the proposed method mitigated the discretization error introduced during the standard backprojection process (yellow arrows in the third column). Quantitative analysis revealed that the proposed method achieved the lowest MAE compared with the FBP and MBHC methods.

We further investigated the performance of the proposed method for the photon starvation effect. The relationship described in \eref{linear-approximation3} is valid when sufficient X-ray photons reach the detector. Assuming that the metal trace of the numerical phantom was significantly affected by photon starvation, we trained the neural network $f_{\Theta}$ in (\ref{MARNeRF}) using a subset of X-rays, denoted by $\mS_t\subseteq \mS$, passing through the teeth and bone only.

Fig. \ref{fig-num-result_ps} compares the reconstruction results of the proposed method trained using the sets $\mS$ (labeled as `w/o photon starvation') and $\mS_t$ (labeled as `w/ photon starvation'). For photon starvation, crown masks were added to the reconstructed image. As indicated by the red arrows, the proposed method trained using $\mS_t$ faced challenges in fully restoring the teeth surrounded by crowns owing to limited information available for recovery. However, the proposed method successfully recovered the morphological structures of the teeth near the crowns.

\subsection{Phantom Experiment}

The phantom experiment was conducted using an industrial CBCT scanner equipped with a flat-panel detector (DUKIN, Korea). The resolution phantom containing the three metallic bolts was scanned using a tube voltage of 160 kVp and tube current of 3.0 mAs. A comparison was performed on the sinogram corresponding to the midplane of the CBCT scan. All CBCT images of size $512\times 512$ were reconstructed with a pixel size of $0.2~\text{mm} \times 0.2~\text{mm}$. The MBHC method corrects metal artifacts using \eref{bhc_formula} with the parameter $\kappa=1$. In the proposed method, the estimate $\hat{\text{P}}$ in \eref{linear-approximation5} is computed using a fan-beam projection operator \cite{Feldkamp1984}. 

Fig. \ref{fig-phan_exp} compares the reconstruction results for the experiment phantom. The first row shows CT images reconstructed using FBP, MBHC, and the proposed method. The insets represent enlarged metal regions, thus highlighting the presence of cupping artifacts. The second row shows background images of the resolution phantom. A background mask was generated manually from the FBP image. The MBHC and proposed method reduced the cupping artifacts in the reconstructed image. Compared with the MBHC method, the proposed method more effectively reduced the streaking and shadowing artifacts caused by the three metallic bolts in the reconstructed images while preserving the structures of the resolution phantom (red arrows in the first row). However, as indicated by the yellow arrow, additional artifacts were introduced in the reconstructed image obtained by the proposed method, possibly owing to other causes of metal artifacts, such as scattering. For a quantitative evaluation, MSEs were computed in the background region. The proposed method demonstrated the lowest MSE value.

\section{Discussion and Conclusion}
This study presented an innovative approach for MAR in dental CBCT by harnessing the regularization power of implicit neural representation techniques. The MLP supplementary output, which captures the nonlinear beam-hardening factor stemming from the polychromatic nature of the X-ray beams, is critical in generating high-quality cross-sectional images. By integrating the MLP with a modified Beer-Lambert law and incorporating X-ray casting of point samples, the proposed method effectively mitigates beam-hardening artifacts, substantially enhancing the overall image quality and increasing the clinical relevance of dental CBCT imaging.

Recently, Kim et al. \cite{Kim2022} introduced an implicit neural representation-based approach for CT reconstruction. Their work focused primarily on sparse-view CT reconstruction and did not specifically address the challenging tasks of MAR. Furthermore, their method relied on existing CT reconstruction techniques. By contrast, our study fills this gap by presenting a novel approach for specifically addressing the MAR in dental CBCT, thereby paving the way for improved image quality. Because our approach is in its initial stages, it can be further improved, thus holding the potential to revolutionize the field of low-dose CT reconstruction.

Implicit neural representations offer substantial advantages over traditional grid-based representations, such as pixels and voxels, particularly in solving ill-posed image reconstruction problems. A key advantage is their resolution-independent capability, wherein the representation capacity is determined by the MPLs capacity rather than the grid resolution. MLPs can capture the underlying structure of an image while minimizing redundancy in the representation without sacrificing accuracy or information content.

Our ongoing research aims to enhance the proposed method based on implicit neural representation, focusing on two critical aspects: improving computational time and achieving accurate 3D reconstruction in dental CBCT. To enhance computational efficiency, the implementation of pre-trained parameters can be investigated using transfer learning, specifically leveraging image priors in dental CBCT. Although our experiments have shown promising capabilities for removing metal-induced artifacts, residual artifacts, particularly thread-like structures, were observed around metal objects. This observation indicates a minor discrepancy between the rendered model used in our method and real-world clinical CBCT data. Therefore, our ongoing research focuses on refining our mathematical model to better align it with the intricacies and nuances of clinical CBCT data.




\stop

%% file: definition.tex
\newtheorem{remark}{Remark}[section]

                    {$\blacksquare$\vspace*{7pt}} 

\def\R{{\mathbb R}}

\def\CT{\mathrm{CT}}

\def\f{\frac}
\def\p{\partial}

\def\bb{{\boldsymbol b}}

\def\bd{{\mathbf d}}

\def\bi{{\mathbf i}}

\def\br{\boldsymbol{r}}

\def\x{\boldsymbol{x}}

\def\A{{\mathbf A}}

\def\E{\mathbf{E}}

\def\W{{\mathbf W}}

\def\mR{{\mathcal R}}
\def\mS{{\mathcal S}}

\def\bi{\begin{itemize}} \def\ei{\end{itemize}}
\def\be{\begin{eqnarray*}}
\def\ee{\end{eqnarray*}}

\def\0{{\mathbf 0}}

\newcommand{\beq}{\begin{equation}}
\newcommand{\eeq}{\end{equation}}

\def\eref#1{(\ref{#1})}

\newcommand{\eps}{\varepsilon}

\def\XXint#1#2#3{{\setbox0=\hbox{$#1{#2#3}{\int}$ }
\vcenter{\hbox{$#2#3$ }}\kern-.55\wd0}}

%% file: NeRF-CT.tex

\tikzset{every picture/.style={line width=0.75pt}} 

\begin{tikzpicture}[x=0.75pt,y=0.75pt,yscale=-1,xscale=1]

\draw  [color={rgb, 255:red, 231; green, 179; blue, 179 }  ,draw opacity=1 ][line width=3] [line join = round][line cap = round] (665.67,234.84) .. controls (665.67,234.84) and (665.67,234.84) .. (665.67,234.84) ;
\draw (319.5,207.09) node  {\includegraphics[width=450.75pt,height=162.13pt]{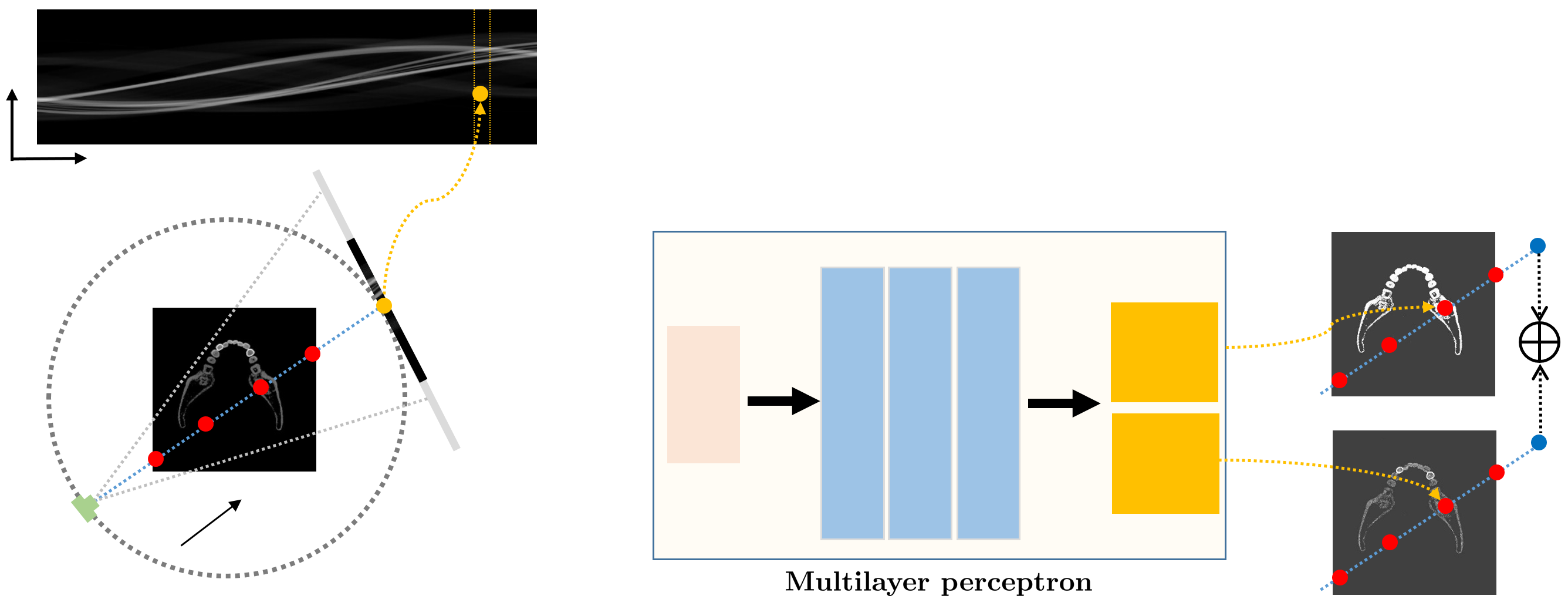}};
\draw [color={rgb, 255:red, 74; green, 144; blue, 226 }  ,draw opacity=1 ][line width=1.5]  [dash pattern={on 1.69pt off 2.76pt}]  (618,221.17) -- (633,221.17) -- (631,115) -- (478,115) -- (478,100) ;
\draw [shift={(476,93)}, rotate = 84] [color={rgb, 255:red, 74; green, 144; blue, 226 }  ,draw opacity=1 ][line width=1.5]    (14.21,-4.28) .. controls (9.04,-1.82) and (4.3,-0.39) .. (0,0) .. controls (4.3,0.39) and (9.04,1.82) .. (14.21,4.28)   ;
\draw [color={rgb, 255:red, 183; green, 170; blue, 20 }  ,draw opacity=1 ][line width=1.5]  [dash pattern={on 1.69pt off 2.76pt}]  (212,132.17) -- (351,131.17) -- (390,100) ;
\draw [shift={(390,100)}, rotate = 140] [color={rgb, 255:red, 183; green, 170; blue, 20 }  ,draw opacity=1 ][line width=1.5]    (14.21,-4.28) .. controls (9.04,-1.82) and (4.3,-0.39) .. (0,0) .. controls (4.3,0.39) and (9.04,1.82) .. (14.21,4.28)   ;
\draw  [color={rgb, 255:red, 184; green, 233; blue, 134 }  ,draw opacity=1 ][fill={rgb, 255:red, 255; green, 255; blue, 255 }  ,fill opacity=1 ] (365,125) -- (653,125) -- (653,175) -- (365,175) -- cycle ;

\draw (230,66) node [anchor=north west][inner sep=0.75pt]   [align=left] {
$
\mbox{Loss }~ \mathcal L =  \frac{1}{|\mathcal{S}|}\displaystyle \sum_{(\varphi, u ,v) \in \mathcal{S}} \left| P( \varphi ,u, v)  - \hat{P}( \varphi ,u,v)  \right|
$};
\draw (370,134) node [anchor=north west][inner sep=0.75pt]  [align=left][font=\footnotesize]  {
$
\hat{P}( \varphi ,u, v) = \int_{0}^{L}\textcolor[rgb]{0.99,0.04,0.04}{\boldsymbol{\mu} }(\boldsymbol{r}( t)) \ dt\ -\ln\left(\frac{\mbox{sinh}\left(\lambda\int _{0}^{L}\textcolor[rgb]{0.98,0.01,0.01}{\boldsymbol{\sigma} }(\boldsymbol{r}( t)) \ dt\right)}{\lambda \int_{0}^{L}\textcolor[rgb]{0.99,0.04,0.04}{\boldsymbol{\sigma} }(\boldsymbol{r}( t)) \ dt}\right) \ \
$};
\draw (360,233.4) node [anchor=north west][inner sep=0.75pt]    {$F_{\Theta }$};
\draw (100,285.4) node [anchor=north west][inner sep=0.75pt]    {$\boldsymbol{d}_{\varphi ,u,v}$};
\draw (283,233.4) node [anchor=north west][inner sep=0.75pt]    {$\boldsymbol{x}$};
\draw (450,255.4) node [anchor=north west][inner sep=0.75pt]    {$\textcolor[rgb]{0.82,0.01,0.11}{\boldsymbol{\sigma} }(\boldsymbol{x})$};
\draw (450,214) node [anchor=north west][inner sep=0.75pt]    {$\textcolor[rgb]{0.82,0.01,0.11}{\boldsymbol{\mu} }(\boldsymbol{x})$};
\draw (29,275.4) node [anchor=north west][inner sep=0.75pt]    {$\boldsymbol{o}_{\varphi }$};
\draw (10,126) node [anchor=north west][inner sep=0.75pt]    {$u$};
\draw (41,158) node [anchor=north west][inner sep=0.75pt]    {$\varphi $};
\draw (101,78) node [anchor=north west][inner sep=0.75pt]    {$P( \varphi ,u,v)$};

\end{tikzpicture}

%% file: MAR-CT.tex
%
%
%

\tikzset{every picture/.style={line width=0.75pt}} 

	\begin{tikzpicture}[x=0.75pt,y=0.75pt,yscale=-1,xscale=1]
		
		\draw (318,155) node  {\includegraphics[width=433.35pt,height=228.75pt]{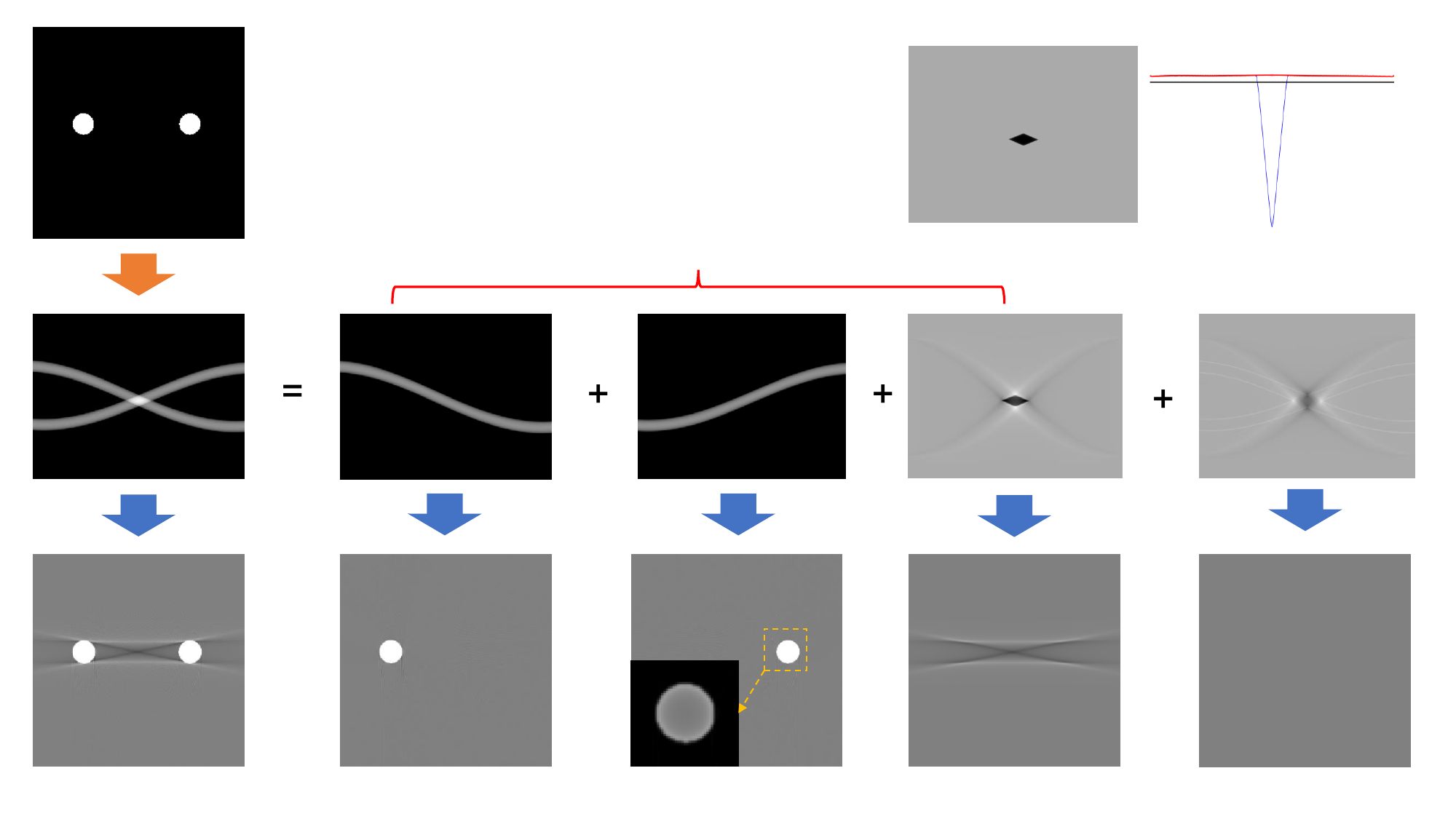}};
		\draw [color={rgb, 255:red, 155; green, 155; blue, 155 }  ,draw opacity=1 ]   (490,92) -- (590,92) ;
	\draw [shift={(590,92)}, rotate = 180] [color={rgb, 255:red, 155; green, 155; blue, 155 }  ,draw opacity=1 ][line width=0.75]    (10.93,-3.29) .. controls (6.95,-1.4) and (3.31,-0.3) .. (0,0) .. controls (3.31,0.3) and (6.95,1.4) .. (10.93,3.29)   ;
	\draw [color={rgb, 255:red, 128; green, 230; blue, 15 }  ,draw opacity=1 ] [dash pattern={on 0.84pt off 2.51pt}]  (527,30) -- (527,92) ;
	\draw [shift={(527,92)}, rotate = 89.17] [color={rgb, 255:red, 128; green, 230; blue, 15 }  ,draw opacity=1 ][fill={rgb, 255:red, 128; green, 230; blue, 15 }  ,fill opacity=1 ][line width=0.75]      (0, 0) circle [x radius= 1.34, y radius= 1.34]   ;
	\draw [color={rgb, 255:red, 128; green, 230; blue, 15 }  ,draw opacity=1 ] [dash pattern={on 0.84pt off 2.51pt}]  (540,30) -- (540,92) ;
	\draw [shift={(540,92)}, rotate = 89.17] [color={rgb, 255:red, 128; green, 230; blue, 15 }  ,draw opacity=1 ][fill={rgb, 255:red, 128; green, 230; blue, 15 }  ,fill opacity=1 ][line width=0.75]      (0, 0) circle [x radius= 1.34, y radius= 1.34]   ;
	\draw [color={rgb, 255:red, 128; green, 230; blue, 15 }  ,draw opacity=1 ] [dash pattern={on 0.84pt off 2.51pt}]  (88,120) -- (88,180) ;
	\draw [shift={(88,180)}, rotate = 88.83] [color={rgb, 255:red, 128; green, 230; blue, 15 }  ,draw opacity=1 ][fill={rgb, 255:red, 128; green, 230; blue, 15 }  ,fill opacity=1 ][line width=0.75]      (0, 0) circle [x radius= 1.34, y radius= 1.34]   ;
	\draw [color={rgb, 255:red, 128; green, 230; blue, 15 }  ,draw opacity=1 ] [dash pattern={on 0.84pt off 2.51pt}]  (80,120) -- (80,180) ;
	\draw [shift={(80,180)}, rotate = 88.83] [color={rgb, 255:red, 128; green, 230; blue, 15 }  ,draw opacity=1 ][fill={rgb, 255:red, 128; green, 230; blue, 15 }  ,fill opacity=1 ][line width=0.75]      (0, 0) circle [x radius= 1.34, y radius= 1.34]   ;
	\draw [color={rgb, 255:red, 81; green, 141; blue, 15 }  ,draw opacity=1 ] [dash pattern={on 0.84pt off 2.51pt}]  (40,39.5) -- (132.62,57.1) -- (137.05,57.94) ;
	\draw [shift={(140,58.5)}, rotate = 190.76] [fill={rgb, 255:red, 81; green, 141; blue, 15 }  ,fill opacity=1 ][line width=0.08]  [draw opacity=0] (10.72,-5.15) -- (0,0) -- (10.72,5.15) -- (7.12,0) -- cycle    ;
	\draw [color={rgb, 255:red, 64; green, 114; blue, 7 }  ,draw opacity=1 ] [dash pattern={on 0.84pt off 2.51pt}]  (40,58) -- (137.06,38.59) ;
	\draw [shift={(140,38)}, rotate = 168.69] [fill={rgb, 255:red, 64; green, 114; blue, 7 }  ,fill opacity=1 ][line width=0.08]  [draw opacity=0] (10.72,-5.15) -- (0,0) -- (10.72,5.15) -- (7.12,0) -- cycle    ;

	\draw [color={rgb, 255:red, 128; green, 230; blue, 15 }  ,draw opacity=1 ] [dash pattern={on 0.84pt off 2.51pt}]  (429,20) -- (429,86) ;
	\draw [shift={(429,86)}, rotate = 89.17] [color={rgb, 255:red, 128; green, 230; blue, 15 }  ,draw opacity=1 ][fill={rgb, 255:red, 128; green, 230; blue, 15 }  ,fill opacity=1 ][line width=0.75]      (0, 0) circle [x radius= 1.34, y radius= 1.34]   ;
	\draw [color={rgb, 255:red, 128; green, 230; blue, 15 }  ,draw opacity=1 ] [dash pattern={on 0.84pt off 2.51pt}]  (441,20) -- (441,86) ;
	\draw [shift={(441,86)}, rotate = 89.17] [color={rgb, 255:red, 128; green, 230; blue, 15 }  ,draw opacity=1 ][fill={rgb, 255:red, 128; green, 230; blue, 15 }  ,fill opacity=1 ][line width=0.75]      (0, 0) circle [x radius= 1.34, y radius= 1.34]   ;
	
	\draw (48,21.4) node [anchor=north west][inner sep=0.75pt]  [color={rgb, 255:red, 248; green, 231; blue, 28 }  ,opacity=1 ]  {$D_{1}$};
	\draw (94,20.4) node [anchor=north west][inner sep=0.75pt]  [color={rgb, 255:red, 248; green, 231; blue, 28 }  ,opacity=1 ]  {$D_{2}$};
	\draw (195,121.4) node [anchor=north west][inner sep=0.75pt]  [color={rgb, 255:red, 248; green, 231; blue, 28 }  ,opacity=1 ]  {$\text{P}_{D_{1}} \ $};
	\draw (76,121.4) node [anchor=north west][inner sep=0.75pt]  [color={rgb, 255:red, 248; green, 231; blue, 28 }  ,opacity=1 ]  {$\text{P}$};
	\draw (309,121.4) node [anchor=north west][inner sep=0.75pt]  [color={rgb, 255:red, 248; green, 231; blue, 28 }  ,opacity=1 ]  {$\text{P}_{D_{2}} \ $};
	\draw (292,83.4) node [anchor=north west][inner sep=0.75pt]  [color={rgb, 255:red, 208; green, 2; blue, 27 }  ,opacity=1 ]  {$\text{P}^{sino}$};
	\draw (538,100.4) node [anchor=north west][inner sep=0.75pt]  [color={rgb, 255:red, 208; green, 2; blue, 27 }  ,opacity=1 ]  {$\text{P}^{\perp }$};
	\draw (58,212.4) node [anchor=north west][inner sep=0.75pt]  [color={rgb, 255:red, 248; green, 231; blue, 28 }  ,opacity=1 ]  {$\mathcal{R}^{-1}( P)$};
	\draw (173,212.4) node [anchor=north west][inner sep=0.75pt]  [color={rgb, 255:red, 248; green, 231; blue, 28 }  ,opacity=1 ]  {$\mathcal{R}^{-1}( \text{P}_{D_{1}}) \ $};
	\draw (288,212.4) node [anchor=north west][inner sep=0.75pt]  [color={rgb, 255:red, 248; green, 231; blue, 28 }  ,opacity=1 ]  {$\mathcal{R}^{-1}( \text{P}_{D_{2}}) \ $};
	\draw (375,121.4) node [anchor=north west][inner sep=0.75pt]  [font=\normalsize,color={rgb, 255:red, 0; green, 0; blue, 0 }  ,opacity=1 ]  {$\text{P}^{sino} -\text{P}_{D_{1}} -\text{P}_{D_{2}} \ $};
	\draw (513,212.4) node [anchor=north west][inner sep=0.75pt]  [color={rgb, 255:red, 248; green, 231; blue, 28 }  ,opacity=1 ]  {$\mathcal{R}^{-1}\left( \text{P}^{\perp }\right)$};
	\draw (260,45.4) node [anchor=north west][inner sep=0.75pt]    {$\text{P}-\text{P}_{D_{1}} -\text{P}_{D_{2}} =\ $};
	\draw (491,47.4) node [anchor=north west][inner sep=0.75pt]  [font=\scriptsize,color={rgb, 255:red, 74; green, 144; blue, 226 }  ,opacity=1 ]  {$\int \text{P}\ ( \varphi ,u) \ du\ $};
	\draw (491,10) node [anchor=north west][inner sep=0.75pt]  [font=\scriptsize,color={rgb, 255:red, 208; green, 2; blue, 27 }  ,opacity=1 ]  {$\int \text{P}^{sino} \ ( \varphi ,u) \ du\ $};
	\draw (580,72.4) node [anchor=north west][inner sep=0.75pt]   [font=\scriptsize] {$\varphi $};
	\draw (540,74.4) node [anchor=north west][inner sep=0.75pt]  [font=\tiny]  {$\frac{5\pi }{9}$};
	\draw (512,74.4) node [anchor=north west][inner sep=0.75pt]  [font=\tiny]  {$\frac{4\pi }{9}$};
	\draw (141,25.4) node [anchor=north west][inner sep=0.75pt]  [font=\tiny]  {$\varphi =\frac{5\pi }{9}$};
	\draw (141,55.39) node [anchor=north west][inner sep=0.75pt]  [font=\tiny,rotate=-359.67]  {$\varphi = \frac{4\pi }{9}$};
\end{tikzpicture}
